\documentclass[]{bytedance_seed}

\usepackage[toc,page,header]{appendix}
\usepackage{hyperref}
\usepackage{url}
\usepackage{graphicx}
\usepackage{tcolorbox}
\tcbuselibrary{breakable}
\usepackage{xspace}
\usepackage{pifont}
\usepackage{makecell}
\usepackage{afterpage}
\usepackage{diagbox}
\usepackage{multicol}

\usepackage{minitoc}

\newcommand{\name}{\text{LPFQA}\xspace}%
\title{\name: A Long-Tail Professional Forum-based Benchmark for LLM Evaluation}

\affiliation[]{ByteDance Seed \quad \quad   Peking University}
\contribution{Full author list in Contributions}

\abstract{
Large Language Models (LLMs) perform well on standard reasoning and question-answering benchmarks, yet such evaluations often fail to capture their ability to handle long-tail, expertise-intensive knowledge in real-world professional scenarios.
We introduce \name, a long-tail knowledge benchmark derived from authentic professional forum discussions, covering 7 academic and industrial domains with 430 curated tasks grounded in practical expertise.
\name evaluates specialized reasoning, domain-specific terminology understanding, and contextual interpretation, and adopts a hierarchical difficulty structure to ensure semantic clarity and uniquely identifiable answers.
Experiments on over multiple mainstream LLMs reveal substantial performance gaps, particularly on tasks requiring deep domain reasoning, exposing limitations overlooked by existing benchmarks.
Overall, \name provides an authentic and discriminative evaluation framework that complements prior benchmarks and informs future LLM development. 
}

\date{\today}
\correspondence{Tong Yang at \email{yangtong@pku.edu.cn}, Zaiyuan Wang at \email{wangzaiyuan@bytedance.com}, Ge Zhang at \email{gezhang@umich.edu}}

\checkdata[Project Page]{\url{https://huggingface.co/datasets/m-a-p/LPFQA}\\
}

\begin{document}
\maketitle


\section{Introduction}
Large Language Models (LLMs) have achieved remarkable progress across a wide range of tasks, including question answering \cite{zhuang2023toolqa,li2024flexkbqa}, logical reasoning \cite{havrilla2024glore,wang2023can}, and code optimization \cite{nam2024using,gu2023llm}, enabling their rapid deployment in both professional applications and everyday use \cite{yang2024talk2care,zheng2025large}. 
As LLM architectures and training paradigms continue to evolve, accurately and comprehensively evaluating their real capabilities has become increasingly important. 
To this end, benchmark-based evaluation remains the primary approach, typically assessing a model’s ability to retrieve and apply knowledge through curated sets of tasks and questions \cite{chang2024survey}.

\begin{figure}[t]
    \centering
    \includegraphics[width=0.7\linewidth]{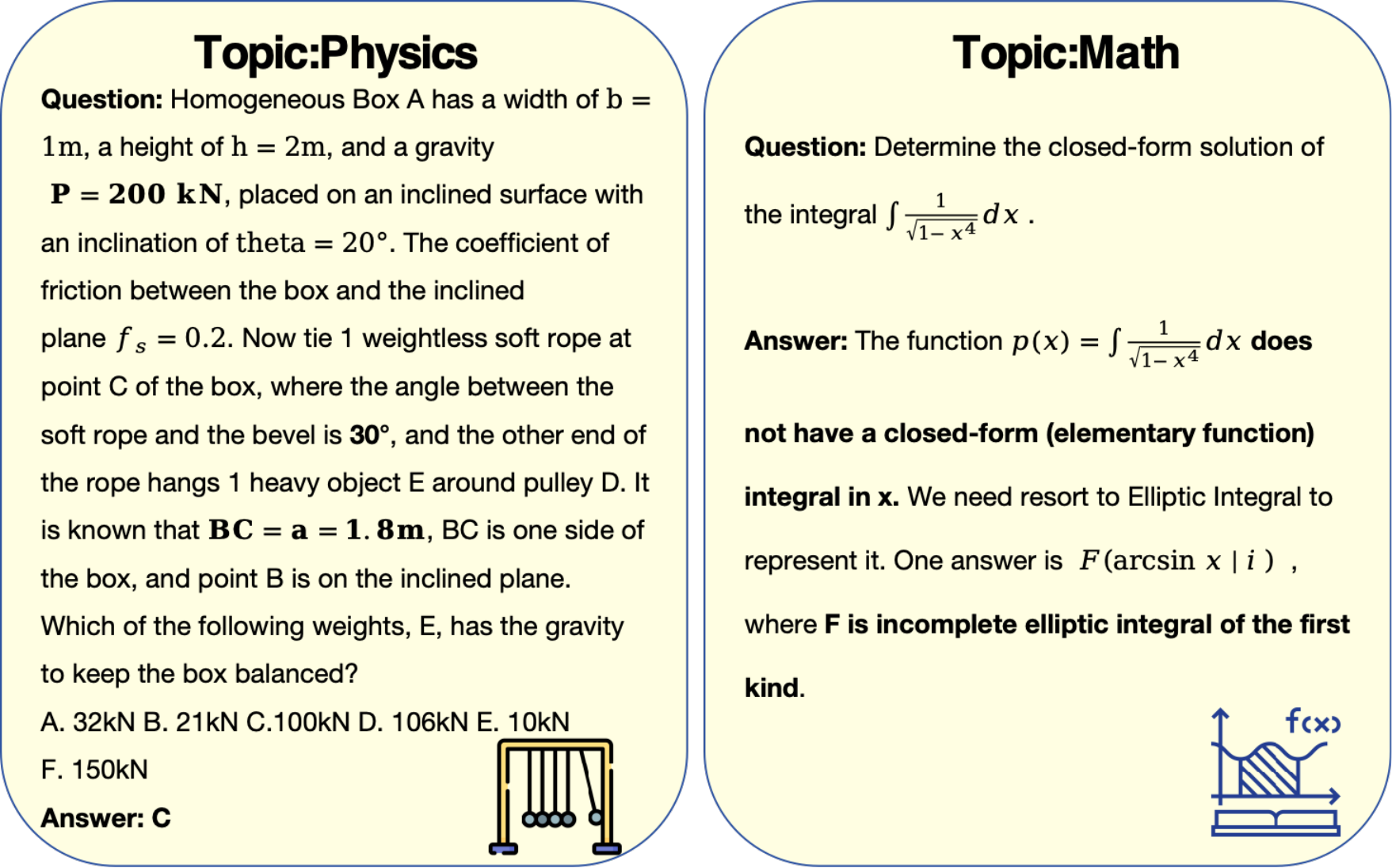}
    \caption{Demos of tasks in \name. The tasks contain the format of multiple choices and short answer, spanning multiple professional domains.}
    \label{fig:example}
\end{figure}

A critical factor in evaluating LLMs is the frequency with which knowledge appears during training \cite{kandpal2023large}.
However, real-world knowledge exhibits a pronounced long-tail distribution, where a small subset of facts is highly frequent while the majority occur sparsely \cite{li2024role,godbole2023benchmarking}.
Highly specialized domain knowledge therefore remains underrepresented in mainstream corpora despite its practical importance.
Benchmarks dominated by high-frequency knowledge consequently fail to reflect a model’s true capabilities \cite{naik2022adapting}.
In contrast, robust generalization, retrieval, and complex reasoning beyond training distributions—abilities essential for real-world deployment—are more faithfully evaluated through performance on long-tail tasks \cite{zhang2023deep,yang2022survey}.

Despite their widespread use, existing benchmarks exhibit systematic limitations in capturing such long-tail capabilities.
Even multi-domain benchmarks remain largely skewed toward frequent or popular knowledge \cite{hendrycksmeasuring,wang2024mmlu}, while others rely on automated or template-based generation that lacks real-world authenticity \cite{mallen2023popqa,kumar2024automatic,huang2025ltgen,li2024crowdsourced,phan2025humanity}.
Moreover, most benchmarks emphasize single-hop retrieval or popularity-driven metrics, overlooking professional-level domain depth, temporal knowledge evolution, and expert validation \cite{mallen2023popqa,zhang2024retrievalqa,sciavolino2021entity}.
As a result, existing evaluations remain insufficient for assessing LLM performance on realistic long-tail professional knowledge.

To this end, we construct a Long-tail Professional Forum QA evaluation benchmark (\name), grounded in discussions from authentic professional forums where real users seek assistance for practical, real-world problems.
Unlike common questions that can be easily resolved through web search or encyclopedic resources, questions raised in such forums typically emerge from real usage scenarios, exhibiting high domain specificity and extremely low frequency in mainstream data sources, thereby naturally aligning with the notion of long-tail knowledge.
As a result, these questions provide a faithful reflection of genuine user demands and the complexity encountered in real-world applications, rather than synthetic or academically curated problem settings.
Through a three-stage data construction pipeline—comprising data collection and pre-processing, automated question generation with quality control, and expert verification with difficulty calibration—we ensure that all tasks in \name consistently preserve their real-world grounding, professional depth, and verified correctness.
\name consists of 430 question items derived from multiple vetted, publicly accessible professional forums, spanning 7 primary academic domains and 21 secondary domains. 
Demos of \name tasks are shown in Figure~\ref{fig:example}.

Based on systematic evaluations of a broad range of mainstream LLMs on \name, we observe that, despite their strong performance on existing knowledge-oriented benchmarks, current models exhibit substantial performance degradation when confronted with long-tail professional questions. These failures are not isolated, but instead reveal systematic error patterns that expose a clear gap between general knowledge proficiency and the core capabilities required for reliable long-tail professional question answering, including domain-specific knowledge coverage and fine-grained question understanding.

The main contributions of our work are summarized as follows: 
\begin{itemize}
   \item We introduce \name, a benchmark for long-tail professional question answering, grounded in real-world discussions from highly specialized online forums, targeting domain-specific knowledge that is largely underrepresented in standard training corpora.
   
   \item We systematically evaluate a broad range of mainstream LLMs on \name, and show that, despite strong performance on existing knowledge-oriented benchmarks, current models suffer from substantial performance degradation on long-tail professional queries.
   
   \item We present a capability-oriented analysis of model failures, revealing systematic error patterns and highlighting the core abilities required for long-tail professional question answering, which provides insights for future model development.
\end{itemize}

    
    
    
\section{Related Work}
\subsection{Long-Tail Knowledge Benchmarks}
LLM evaluation has rapidly evolved from early benchmarks~\cite{wang2018glue,wang2019superglue}, which focus on general language understanding, to more comprehensive benchmarks like MMLU~\cite{wang2024mmlu}, BIG-bench~\cite{srivastava2023beyond}, and HELM~\cite{liang2022holistic}, which extend evaluation to multi-disciplinary knowledge, reasoning, and holistic dimensions. 
Despite these advances, existing benchmarks remain limited in their ability to capture long-tail knowledge and authentic, user-centric information needs encountered in real-world scenarios, motivating the development of new evaluation paradigms that go beyond curated or high-frequency knowledge distributions.

Given the significance of long-tail knowledge in assessing the authentic performance of LLMs, researchers have developed various specialized benchmarks through two primary construction paths. 
The first involves natural data collection from real-world distributions, such as the iNaturalist biodiversity dataset \cite{van2018inaturalist}. 
The second method utilizes synthetic construction by creating imbalanced sampling from existing datasets, exemplified by ImageNet-LT \cite{liu2019large}. Despite these efforts, current long-tail knowledge benchmarks exhibit several pervasive deficiencies in practical applications. 
First, there is a notable lack of authenticity and naturalness in task design. 
Benchmarks like PopQA \cite{mallen2023popqa} and Automatic QA Generation \cite{kumar2024automatic}, rely on templates or static knowledge graphs like LTGen \cite{huang2025ltgen}. 
This leads to mechanical language patterns and fails to capture the dynamic, evolving nature of real-world information, a limitation that persists even in studies like RetrievalQA \cite{zhang2024retrievalqa} due to their narrow coverage. 
Second, the task complexity is frequently insufficient. Most datasets, including Entity-Questions \cite{sciavolino2021entity}, focus primarily on single-hop factual recall while neglecting complex reasoning and cross-domain inference. 
The reliance on automated pipelines without systematic human verification may result in compromised data quality and factual inaccuracies. 


\subsection{User-Centric and Challenging Benchmarks}

Beyond static long-tail knowledge evaluation, another line of work focuses on user-centric and dynamic assessment of model capabilities.
Chatbot Arena~\cite{chiang2024chatbot} evaluates LLMs through crowdsourced blind comparisons, where users interact with two anonymous models and vote for the better response.
This approach effectively captures user preferences in open-ended conversations.
However, such crowdsourced evaluations provide limited control over task difficulty and domain expertise, and user-submitted queries are often simple, resulting in reduced discriminative power among strong models.
Arena-Hard~\cite{li2024crowdsourced} partially addresses this issue via adversarial questioning, but its question distribution remains relatively concentrated, limiting its ability to assess a broad range of complex professional long-tail scenarios.To push models toward more challenging settings, Humanity’s Last Exam (HLE)~\cite{phan2025humanity} introduces extremely difficult questions designed to test advanced reasoning and cross-disciplinary knowledge.
While HLE serves as a valuable probe of frontier-level capabilities, its extreme difficulty and limited alignment with everyday user needs reduce its suitability as a general-purpose or regularly applied evaluation benchmark.
In general, existing user-centric and extreme-difficulty benchmarks complement long-tail knowledge evaluation but remain insufficient for systematically measuring complex reasoning over professional long-tail knowledge in realistic application settings.
Our work aims to bridge this gap by constructing a benchmark that combines controlled difficulty, domain expertise, and real-world relevance.
\begin{figure*}[t]
    \centering
    \includegraphics[width=\linewidth]{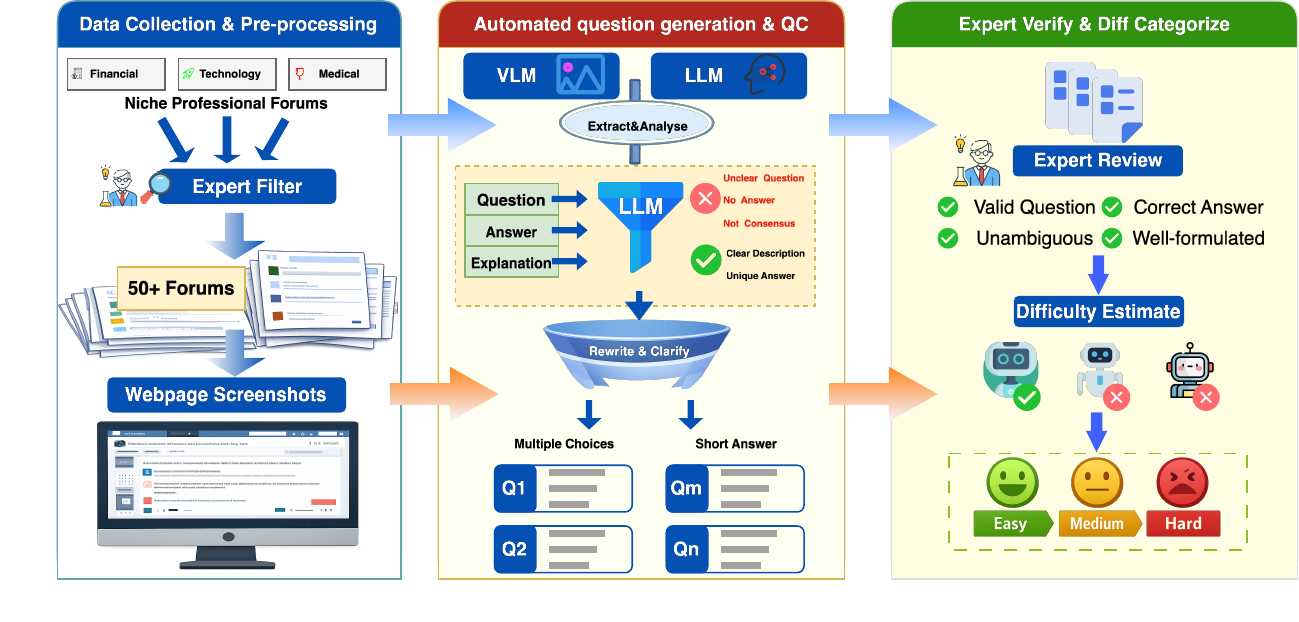}
    \caption{Construction pipeline of \name.}
    \label{fig:pipeline}
\end{figure*}

\section{\name Benchmark}
To systematically evaluate LLM's ability to understand and reason over long-tail professional questions in real-world scenarios, \name is constructed by collecting niche questions posed by users along with their authoritative answers from a wide range of real technical forums and professional platforms across different domains.
Each question is formulated as either a multiple-choice or short-answer problem and is associated with a unique reference answer, ensuring the objectivity and reliability of the evaluation.

 To construct the dataset, we adopt a three-stage pipeline shown in Figure~\ref{fig:pipeline}, consisting of \textbf{data collection\&pre-processing}, \textbf{automated question generation\&quality control}, and \textbf{expert verification\& difficulty categorization}. The details of each stage and information of \name are described in the following subsections.

\subsection{Construction of \name}

\subsubsection{Data Collection and Pre-processing}
In the first stage, we focus on collecting domain-specific discussions which reflect realistic long-tail question-answering scenarios. To this end, we invite human experts to curate professional online forums across diverse specialized domains. For each candidate forum, experts are required to submit its domain classification, access link, and a small set of representative discussion samples. To ensure both domain relevance and data authenticity, submissions are further examined through cross-review among the expert team. This quality control process verifies that the selected forums indeed correspond to specialized professional fields, and that the sampled questions represent realistic, long-tail inquiries arising in real-world usage rather than artificially constructed cases.The detailed qualification statements and rules of recruited experts are presented in Appendix~\ref{appsec:experts_qualification}.

After this filtering process, we employ web crawling tools to collect raw discussion data from over 50 vetted forums, resulting in \textbf{over 2,000} discussion threads within the past 7 months\footnote{This work was conducted in July 2025, that is, all collected discussion threads were posted after January 2025.}.
Given the substantial heterogeneity in webpage structures across different platforms, and in order to preserve as much original contextual information as possible—including layout, formatting, and question–answer relationships—for downstream annotation, all discussions are collected and stored in the form of screenshots rather than plain text. The full list of forums is shown at Appendix~\ref{app:forums}.

\subsubsection{Automated Question Generation and Quality Control}
Building upon the collected visual data, we further leverage  LLMs and vision–language models (VLMs) to perform preliminary filtering and structured annotation. Specifically, each screenshot is first processed by a VLM to extract its content into a structured representation with predefined fields, including the question, candidate answer(s), and accompanying explanations.
Based on the extracted information, an LLM is then employed to assess whether the discussion contains a clearly defined question paired with a valid and unambiguous answer. Cases involving vague problem statements, missing answers, or controversial or incorrect responses are removed at this stage to ensure answer reliability and certainty.

For the remaining tasks, we categorize them according to its nature and assigns an appropriate question format. Knowledge-oriented queries are predominantly transformed into multiple-choice questions, where the LLM constructs one correct option and 4-6 plausible distractors grounded in the original discussion content. In contrast, questions that require reasoning or numerical computation are formulated as short-answer questions, with the model rewriting concise, logically coherent, and well-justified reference answers. Regardless of question type, the model further rewrites and refines the problem descriptions to improve clarity, conciseness, and formal rigor, ensuring that each item is suitable for standardized evaluation. The prompts of models used during this stage can be found in Appendix~\ref{app:llmprompt}. We retain \textbf{approximately 1,000} candidate tasks at the end of this stage.

\begin{table*}[t]
\centering
\resizebox{0.8\linewidth}{!}{
    \begin{tabular}{clccc}
    \toprule
    
    \textbf{Primary Domain} & \textbf{Secondary Domain} & \textbf{Multiple-choice} & \textbf{Short-answer} & \textbf{Total Num} \\
    \midrule
    \multirow[c]{3}{*}{\makecell[c]{Basic Natural Sciences\\(Natural Sci)}} 
     & Physics & \multirow[c]{3}{*}{72} & \multirow[c]{3}{*}{25} &\multirow[c]{3}{*}{\textbf{97}} \\
     & Chemistry &    &&\\
     & Biology &    &&\\
    \midrule
    \multirow[c]{2}{*}{\makecell[c]{Life and Medical Sciences\\(Life Sci)}} 
     & Biology & \multirow[c]{2}{*}{57} & \multirow[c]{2}{*}{14}  &\multirow[c]{2}{*}{\textbf{71}} \\
     & Medical &    &&\\
    \midrule
    \multirow[c]{3}{*}{\makecell[c]{Mathematics and Logic\\(Maths)}} 
     & Mathematics & \multirow[c]{3}{*}{30} & \multirow[c]{3}{*}{31} &\multirow[c]{3}{*}{\textbf{61}} \\
     & AI and Machine Learning & & &\\
     & Data Science & & &\\
    \midrule
    \multirow[c]{2}{*}{\makecell[c]{Computer Science\\(CS)}} 
     & Computer Science & \multirow[c]{2}{*}{9} & \multirow[c]{2}{*}{49} &\multirow[c]{2}{*}{\textbf{58}} \\
     & Computer Systems and Software & & &\\
    \midrule
    \multirow[c]{4}{*}{\makecell[c]{Electronics Engineering\\(EE)}} 
     & Electronic Science and Technology & \multirow[c]{4}{*}{43} & \multirow[c]{4}{*}{19} &\multirow[c]{4}{*}{\textbf{62}} \\
     & Electronics Information Engineering & & &\\
     & Electronics\&Information Science & & &\\
     & Info\&Communication Engineering & & &\\
    \midrule
    \multirow[c]{4}{*}{\makecell[c]{Applied Technologies\\(Applied Tech)}} 
     & General Engineering & \multirow[c]{4}{*}{18} & \multirow[c]{4}{*}{20} &\multirow[c]{4}{*}{\textbf{38}} \\
     & Mechanical and Automation &    &&\\
     & Energy &    &&\\
     & Aerospace &    &&\\
    \midrule
    \multirow[c]{3}{*}{\makecell[c]{Social Sciences\\(Social Sci)}} 
     & Law & \multirow[c]{3}{*}{19} & \multirow[c]{3}{*}{24} &\multirow[c]{3}{*}{\textbf{43}} \\
     & Finance &    &&\\
     & Miscellaneous &    &&\\
    \midrule
    \textbf{Overall} & -- & \textbf{249} & \textbf{181} &\textbf{430} \\
    \bottomrule
    \end{tabular}
}
\caption{Domain distribution of \name benchmark.}
\label{tab:domains}
\end{table*}

\subsubsection{Expert Verification and Difficulty Categorization}
\label{subsec:3.1.3}
The remaining candidate tasks are subsequently subjected to expert-driven refinement and difficulty calibration. 

\textbf{Expert manual verification:} 
Domain specialists carefully review each item to verify that it satisfies the long-tail criterion and that its question–answer pair is unambiguous, well-formulated, and factually correct. Any task failing to meet either requirement is discarded to ensure both realism and answer reliability.

\textbf{Difficulty categorize by preliminary experiment:} 
For the expert-approved tasks, we further conduct difficulty stratification through a pilot evaluation protocol. Specifically, each task is posed to a diverse set of state-of-the-art LLMs, and its difficulty level is determined based on the models’ answer accuracy. Following this procedure, tasks are categorized into three difficulty tiers (\textit{i.e.}, easy, medium, and hard) and overly trivial items are removed from the dataset. 
The detailed criteria for difficulty partitioning are provided in Appendix \ref{appsec:difficulty_distribution}.


\subsection{Statistics of \name} 
We construct \name, a benchmark comprising \textbf{430} professional questions, including 249 multiple-choice items and 181 short-answer questions, spanning 7 primary domains and 21 secondary domains, as summarized in Table~\ref{tab:domains}.
Each domain comprises authentic, highly specialized, and long-tail questions derived from real-world scenarios, paired with expert-verified reference answers. Typical examples of tasks from different domains can be found in Appendix~\ref{app:case}.
\name is designed to rigorously evaluate LLMs not only in terms of their domain-specific factual knowledge in advanced and niche areas, but also their reasoning, analytical, and computational capabilities when addressing complex, professional-level queries. 
As described in \S\ref{subsec:3.1.3}, each question in \name is annotated with a difficulty level. 
The distribution of questions across different difficulty tiers is presented in Appendix~\ref{appsec:difficulty_distribution}.

\begin{table*}[t]
\centering
\setlength{\tabcolsep}{6pt}
\resizebox{\linewidth}{!}{
\begin{tabular}{lcccccccc}
\toprule
\textbf{Model} 
& \textbf{Life Sci} 
& \textbf{Maths} 
& \textbf{Natural Sci} 
& \textbf{Social Sci} 
& \textbf{EE} 
& \textbf{Applied Tech} 
& \textbf{CS} 
& \textbf{Overall} \\
\midrule
Seed-1.6-Thinking
& 41.67 & 52.46 & 46.88 & 41.86 & 42.37 & \textbf{57.89} & 37.93 & 45.35 \\

GLM-4.6
& 54.17 & 55.74 & 45.83 & 48.84 & \textbf{64.41} & 39.47 & 43.10 & 50.47 \\

Gemini-2.5-Pro
& 47.22 & 55.74 & 45.83 & 46.51 & 59.32 & 31.58 & 43.10 & 47.67 \\

Gemini-3-Pro
& 45.83 & 47.54 & 55.21 & 48.84 & 57.63 & \textbf{57.89} & \textbf{48.28} & 51.16 \\

Kimi-K2-Thinking
& 54.17 & 47.54 & 47.92 & 53.49 & 54.24 & 36.84 & 39.66 & 47.91 \\

GPT-5-High
& 45.83 & \textbf{63.93} & \textbf{57.29} & \textbf{55.81} & 61.02 & 52.63 & 44.83 & \textbf{54.42} \\

Claude-Sonnet-4.5-Thinking
& \textbf{56.94} & 49.18 & 50.00 & 44.19 & 57.63 & 42.11 & 43.10 & 49.53 \\

GPT-4.1
& 45.83 & 54.10 & 46.88 & 41.86 & 57.63 & 34.21 & 41.38 & 46.51 \\

DeepSeek-V3.2-Thinking
& 54.17 & 57.38 & 48.96 & 46.51 & 50.85 & 47.37 & 41.38 & 49.53 \\

Qwen-3-235B-Instruct & 43.48 & 60.00 & 47.37 & 46.51 & 46.55 & \textbf{57.89} & 44.83 & 49.09 \\

Qwen3-235B-Thinking
& 40.58 & 48.33 & 52.63 & 48.84 & 51.72 & 36.84 & 46.55 & 46.85 \\

\midrule
\textbf{All Models Avg}
& 48.17 & 53.81 & 49.53 & 47.57 & 54.85 & 44.97 & 41.26 & 48.96 \\
\bottomrule
\end{tabular}
}
\caption{Model performance on \name across primary domains. The best score in each domain is highlighted in bold.}
\label{tab:mainresult_score}
\end{table*}

\section{Experiments}

Based on \name, we evaluate a diverse set of 11 representative LLMs, covering both closed-source and open-source systems: 
5 open-source models include 
DeepSeek-V3.2-Thinking~\cite{deepseekv32}, 
Kimi-K2-Thinking~\cite{team2025kimi}, 
GLM-4.6~\cite{glm4-6}, 
Qwen-3-235B-Instruct and Qwen3-235B-Thinking~\cite{yang2025qwen3};
6 closed-source models include 
GPT-4.1~\cite{openai2024gpt41}, 
GPT-5-high~\cite{GPT5_2025}, 
Gemini-2.5-Pro~\cite{comanici2025gemini}, 
Gemini-3.0-Pro~\cite{gemini3}, 
Claude-Sonnet-4.5-Thinking~\cite{claude4-5}, 
, and Seed-1.6-Thinking~\cite{volcengine2024doubao}. 
To improve the reliability of the evaluation, we set $temperature = 1 $ and $top_p = 1$ for all models in the experiment.

For automatic evaluation, we employ an LLM-based answer verification protocol.
For each task, the question, expert-verified reference answer, and model-generated output are jointly provided to a judge model.
The judge is strictly used to verify whether the final answer matches or is semantically equivalent to the reference answer, under a binary scoring scheme (1 for correct, 0 otherwise).
The final score of each model is computed as the average binary accuracy over all tasks. The prompt of LLM judge can be found in Appendix~\ref{app:llmprompt}.

\subsection{Main Results}
As shown in Table \ref{tab:mainresult_score}, the performance of different models across domains reveals several important observations.

\textbf{Current models remain far from adequately solving real-world long-tail question answering tasks}. 
The average scores of all evaluated models fall within a narrow range of 45\%–55\%. Even the best-performing model, GPT-5, achieves only 54.42\%, which still fails to reach the commonly accepted passing threshold of 60\%.
Given that all questions in \name are collected from real user queries posted on online forums, this result highlights a critical gap between current LLM capabilities and the demands of authentic long-tail problem solving. In particular, it suggests that existing models struggle to reliably handle domain-specific, low-frequency questions that require precise understanding and accurate reasoning beyond memorized patterns. This further underscores the necessity of evaluating LLMs under realistic, user-driven long-tail settings, rather than relying solely on curated or distributionally simplified benchmarks.

\textbf{Current models exhibit substantial performance variation across domains}, indicating that real-world long-tail question answering remains highly domain-dependent.
As shown in Table \ref{tab:mainresult_score}, average model accuracy differs markedly across domains. Mathematics and Electrical Engineering achieve relatively higher average scores (above 52\%), whereas Computer Science and Applied Technology present the lowest overall performance, with average accuracy dropping to approximately 40–45\%. Life Sciences, Natural Sciences, and Social Sciences fall between these extremes but still remain far from reliable. This uneven performance distribution suggests that \name captures heterogeneous domain difficulty, reflecting the diverse nature of real-world long-tail questions rather than a uniformly challenging benchmark.

\textbf{No existing model demonstrates robust and consistent dominance across all domains}, highlighting the limited cross-domain generalization of current LLMs under long-tail settings.
Although GPT-5-High achieves the highest overall score, its performance varies significantly across fields, with notable drops in Life Sciences and Computer Science. Similarly, Claude-Sonnet-4.5 (Thinking) excels in Life Sciences, while Gemini-3-Pro shows relative strength in Computer Science, yet neither maintains stable superiority across domains. This indicates that performance gains are often domain-specific and suggests that scaling model capacity alone is insufficient to ensure reliable performance on diverse, forum-derived long-tail questions.

\subsection{Analysis and Discussions}
\subsubsection{Error Attribution in Long-Tail Professional Tasks
}

\begin{figure*}[t]
    \centering
    \begin{minipage}{0.45\linewidth}
        \centering
        \includegraphics[width=\linewidth]{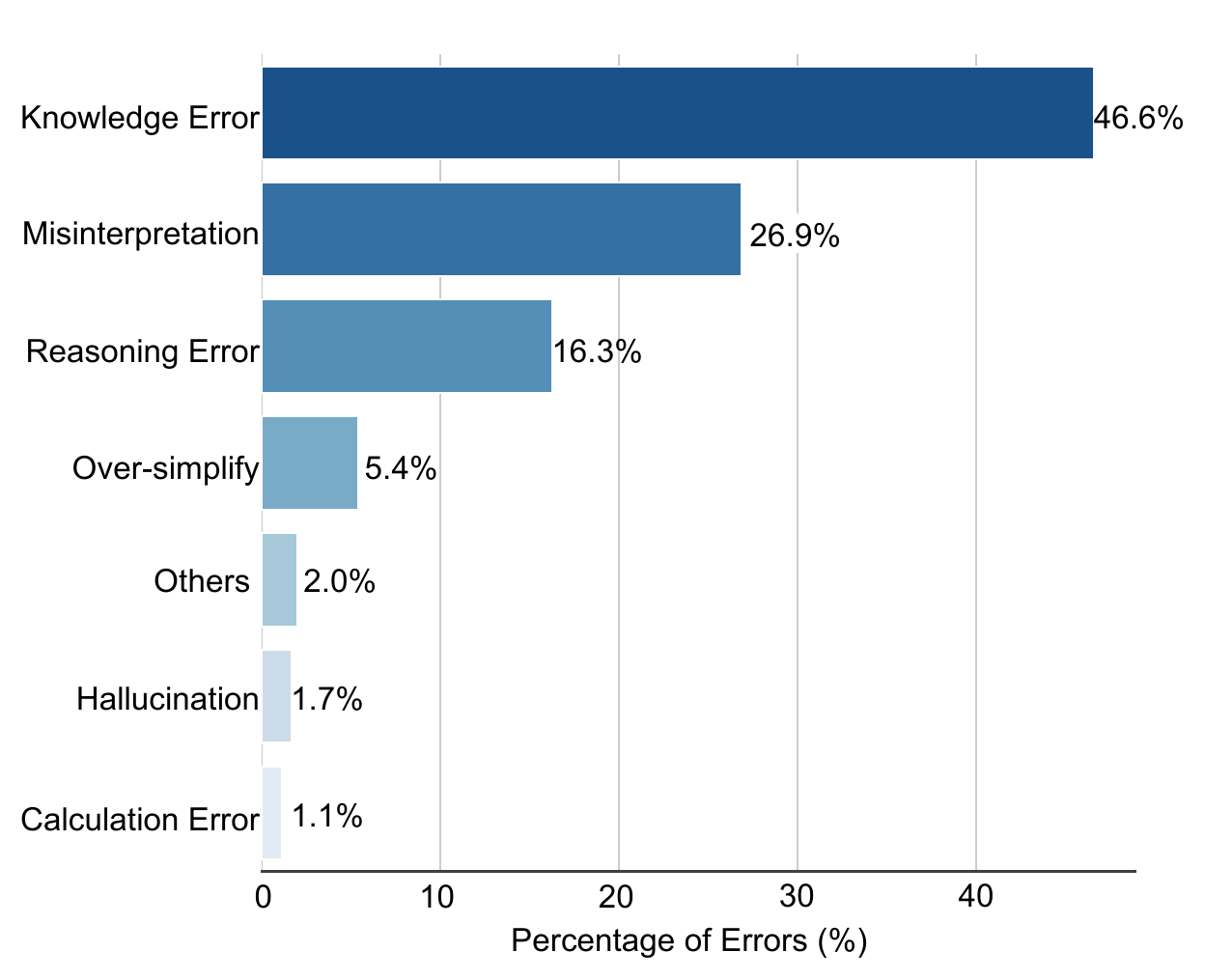}
        \caption{Distributions of error types in the experiment.}
        \label{fig:errordistribution}
    \end{minipage}
    \hfill
    \begin{minipage}{0.45\linewidth}
        \centering
        \includegraphics[width=\linewidth]{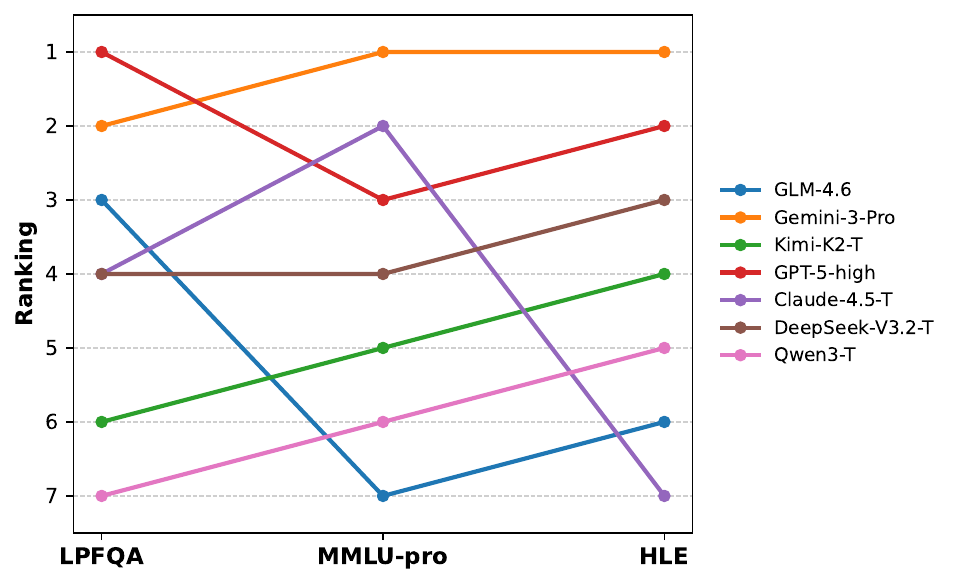}
        \caption{Comparison of model performance between \name and other knowledge-oriented benchmarks.}
        \label{fig:lpfqavsothers}
    \end{minipage}
\end{figure*}

To provide actionable insights for improving model performance on professional long-tail question answering, we conduct a fine-grained error attribution analysis. Specifically, for each primary domain, we randomly sample 50 instances that are incorrectly answered by a model, and systematically categorize the underlying causes of these errors. As illustrated in Figure \ref{fig:errordistribution}, \textbf{knowledge deficiencies and question misinterpretation together account for over 73\% of all observed errors}, indicating that current models still struggle to reliably internalize and correctly apply specialized domain knowledge, as well as to accurately comprehend nuanced and technically phrased questions in long-tail professional scenarios.

In addition, reasoning-related errors constitute 16.3\% of the total error cases, further suggesting that, in high-difficulty and highly specialized settings—particularly within science and engineering domains—correct model responses often require not only factual knowledge but also non-trivial reasoning capabilities. This highlights the persistent gap between existing LLMs and the level of structured reasoning needed to solve complex real-world professional problems. The detailed distribution of error types for each primary domain can be found in Appendix~\ref{app:primary error attribute}.

\subsubsection{Assessing Pre-training Exposure in \name}

To assess the extent to which \name questions may have been exposed during model pre-training, we measure several open-source models on \name and widely used evaluation benchmarks, including MMLU-devset \cite{hendrycksmeasuring} and Math500 \cite{mayilvahanan2025math}, using ROUGE-L~\cite{lin2004rouge} as a lexical overlap metric.
As shown in Table \ref{tab:contamination}, \name exhibits substantially lower ROUGE-L scores compared to standard benchmarks across all 3 models.
In particular, the ROUGE-L scores for \name remain notably below those of MMLU-devset and Math500, indicating limited surface-level overlap with data patterns commonly observed in existing benchmarks.

These results suggest that \name tasks are unlikely to have been extensively encountered during model pre-training.
The relatively low ROUGE-L score reflects the fact that \name queries originate from real-world long-tail professional scenarios rather than curated academic-style benchmarks.
This also indicates that \name occupies a distinct region of the data distribution that is underrepresented in commonly used training and evaluation corpora, and is therefore unlikely to suffer from severe benchmark memorization or data leakage effects. Overall, our analysis provides empirical evidence that \name exhibits low pre-training overlap, supporting its characterization as a benchmark that captures genuinely long-tail question distributions and reinforcing its suitability for evaluating LLM performance on realistic, low-frequency professional QA tasks.

\begin{table}[t]
\centering
    \begin{tabular}{lccc}
    \toprule
    \textbf{Benchmark} & \textbf{Qwen3-30B-Base} & \textbf{Qwen3-30B-Thinking} & \textbf{GPT-oss-20B} \\
    \midrule
    \name          & 24.02 & 33.73 & 25.56\\
    MMLU-devset    & 40.76 & 46.17 & 38.05 \\
    Math500        & 61.35 & 52.47 & 35.84 \\
    \bottomrule
    \end{tabular}
\caption{ROUGE-L scores on different benchmarks. We set the input proportion of input problems to 40\% for all benchmarks. Models are deployed locally in pre-train mode without chat template.}
\label{tab:contamination}
\end{table}

\subsubsection{Relationship of Performance between \name and Existing Knowledge QA Benchmarks}

\definecolor{darkred}{HTML}{CD2626}
\definecolor{darkgreen}{HTML}{228B22}

\begin{table*}[t]
    \centering
    \begin{tabular}{c | c | cc | cc}
        \toprule
        \multirow{2}{*}{\makecell{\diagbox{\textbf{Models}}{\textbf{Tools}}}} 
        & \textbf{No tools}
        & \multicolumn{2}{c|}{\textbf{Jupyter CI}} 
        & \multicolumn{2}{c}{\textbf{Search}} \\
        \cmidrule(lr){2-2} \cmidrule(lr){3-4} \cmidrule(lr){5-6}
        & \textbf{Score}
        & \textbf{Score} & $\Delta$ 
        & \textbf{Score} & $\Delta$ \\
        \midrule
        GPT-5-High 
        & 54.42
        & 44.50 & 18.20\%\textcolor{darkred}{$\downarrow$} 
        & 44.30 & 18.57\%\textcolor{darkred}{$\downarrow$} \\
        
        Gemini-2.5-Pro 
        & 47.67
        & 33.50 & 30.06\%\textcolor{darkred}{$\downarrow$} 
        & 30.50 & 36.33\%\textcolor{darkred}{$\downarrow$} \\
        
        Deepseek-V3.2-Thinking
        & 49.53
        & 31.78 & 32.95\%\textcolor{darkred}{$\downarrow$} 
        & 31.78 & 32.95\%\textcolor{darkred}{$\downarrow$} \\

        Kimi-K2-Thinking 
        & 47.91
        & 33.50 & 30.06\%\textcolor{darkred}{$\downarrow$} 
        & 30.50 & 36.33\%\textcolor{darkred}{$\downarrow$} \\

        GLM-4.6 
        & 50.47
        & 34.80 & 30.95\%\textcolor{darkred}{$\downarrow$} 
        & 37.30 & 25.99\%\textcolor{darkred}{$\downarrow$} \\
        
        \midrule
        \textbf{Average} 
        & 50.00
        & 35.62 & 28.76\%\textcolor{darkred}{$\downarrow$} 
        & 34.87 & 30.26\%\textcolor{darkred}{$\downarrow$} \\
        \bottomrule
    \end{tabular}
    \caption{Performances of tool-equipped models. ``No tools'' denotes the baseline setting without any external tools, while $\Delta$ indicates the relative change compared to this baseline.}
    \label{tab:tool-equipped}
\end{table*}

To investigate the relationship between model performance on professional long-tail tasks and existing knowledge-oriented benchmarks, we compare both the absolute performance and relative rankings of representative models across \name, MMLU-pro~\cite{wang2024mmlu}, and HLE~\cite{phan2025humanity}, as shown in Figure \ref{fig:lpfqavsothers}. Top-models like GPT-5-high and Gemini-3-Pro, consistently remain within the top tier across all three benchmarks, suggesting a partial alignment between LPFQA and conventional knowledge-focused evaluations. This indicates that general factual competence and broad-domain reasoning ability still play an important role in professional long-tail tasks.

However, the figure also reveals substantial ranking divergences between LPFQA and traditional knowledge benchmarks. For example, Gemini-3-Pro, which ranks first on both MMLU-pro and HLE, is surpassed by GPT-5-high on LPFQA. Conversely, Claude-4.5-thinking  demonstrate competitive performance on MMLU-pro but exhibit notable degradation on HLE and LPFQA. These discrepancies suggest that high performance on knowledge-intensive benchmarks does not necessarily translate to strong performance in professional long-tail scenarios. Overall, LPFQA captures a complementary ability dimension that is only partially reflected by existing knowledge and exploration-oriented benchmarks. This further motivates the necessity of evaluating models \textbf{under realistic, domain-specific long-tail settings} beyond conventional knowledge-centric benchmarks. The detailed scores of models on MMLU-pro and HLE can be found in Appendix~\ref{app:scoreonotherbench}.

\subsection{Ablation Experiments: Impact of Tool Augmentation}

To investigate whether external tool augmentation can influence model performance on \name, we select several models and conduct a set of ablation experiments by equipping models with additional tools during inference. Specifically, we apply two commonly used categories of tools: (i) Jupyter Code Interpreter (CI) that enables executable reasoning through code execution, and (ii) Search tools, including Google Search and Text Browser, which allow models to access external web content. 
As shown in Table~\ref{tab:tool-equipped}, we observe a clear and consistent performance degradation across all models when equipped with external tools, regardless of the specific tool type, compared to their corresponding tool-free variants.

Through detailed case analysis, we find that when models are augmented with the CI tool, they tend to reframe questions as programming problems and attempt to solve them via explicit code execution. However, \name questions are characterized by dense domain knowledge and implicitly grounded, knowledge-driven reasoning, which is often difficult—or even impossible—to formalize in a programming language. As a result, forced code-based reasoning can interfere with correct problem solving rather than facilitate it.
Similarly, when equipped with a search tool, models frequently suffer from misleading or low-quality retrieval results. Due to the niche, long-tail nature of \name questions, relevant and reliable evidence is often not retrievable. Instead, the search tool may return superficially related but ultimately irrelevant or even incorrect information. Such noisy retrieval can distract the model and introduce spurious cues, causing it to produce incorrect answers for questions that it could otherwise solve correctly without tool assistance. 
These findings are consistent with both our intended characterization of \name—its focus on implicit, knowledge-intensive reasoning beyond explicit proceduralization—and prior observations that equipping models with tools misaligned with the task requirements can lead to performance degradation rather than improvement~\cite{liu2025wtu,shi2025retrieval}. Detailed case studies of model performances when with/without tools can be found in Appendix~\ref{app:caseontool}.

\section{Conclusion}

In this work, we present \name, a long-tail professional forum–based benchmark for evaluating LLMs on low-frequency, expertise-intensive tasks grounded in real-world scenarios. Our evaluations reveal consistently weak performance across models, indicating substantial difficulty on \name. At a high level, these limitations stem from insufficient mastery of domain-specific knowledge and contextual understanding, which cannot be reliably compensated by external tools. These results demonstrate that \name explicitly challenges LLMs’ ability to reason with long-tail professional knowledge under realistic conditions. \name exposes fundamental gaps in current models and provides a focused evaluation of their capacity for long-tail professional reasoning.

\clearpage
\section{Contributions and Acknowledgments}
\subsection*{Co-first authors}
Liya Zhu, Peizhuang Cong, Jingzhe Ding, Aowei Ji

\subsection*{Contributors}
Wenya Wu, Jiani Hou, Chunjie Wu, Xiang Gao, Jingkai Liu, Huan Zhou, Xuelei Sun, Yang Yang, Jianpeng Jiao, Liang Hu, Xinjie Chen, Jiashuo Liu

\subsection*{Corresponding Authors}
Tong Yang, Zaiyuan Wang, Ge Zhang

\subsection*{Advisor}
Wenhao Huang

\clearpage
\bibliographystyle{unsrtnat}
\bibliography{main}

\clearpage
\beginappendix

\section{Examples Q\&A of \name}
\label{app:case}

\begin{tcolorbox}[breakable, title={Q\&A 1, Field: \textit{General Engineering}}]
\textbf{Question}:
When only 110V service is available for a Millport milling machine with a 220V single-phase motor, which power supply solution is recommended, and what key factor must be considered for selecting this equipment?

A. A voltage regulator with adjustable output, focusing on maximum current capacity alone.

B. A three-phase to single-phase converter with 110V input, needing to match the motor speed rating.

C. A Variable Frequency Drive (VFD) with single-phase input and three-phase output, requiring matching the motor's horsepower (HP) rating and current requirements.

D. A step-up transformer with single-phase input and single-phase output, requiring matching the voltage ratio only.

E. A capacitor-start motor conversion kit, requiring compatibility with motor phase configuration.

F. A DC power supply with inverter function, needing to match the motor frequency range.

\textbf{Answer}: C
\end{tcolorbox}

\begin{tcolorbox}[breakable, title={Q\&A 2, Field: \textit{Biology}}]
\textbf{Question}:
According to the Wright Fisher model of population genetics, please calculate how many generations of living offspring would an average person typically have before the extinction of their genetic lineage?

A. Approximately 10 generations

B. Approximately 20 generations

C. Approximately 25 generations

D. Approximately 30 generations

E. Approximately 40 generations

F. Approximately 60 generations

\textbf{Answer}: B
\end{tcolorbox}

\begin{tcolorbox}[breakable, title={Q\&A 3, Field: \textit{Physics}}]
\textbf{Question}: Which of the following accurately describes stellar rotation and the key physical factors determining rotation rate and axis orientation?

A. Not all stars rotate, with key determinants including stellar mass, angular momentum loss during evolution, binary interactions, planetary system formation, and magnetic activity effects

B. All stars rotate, with key determinants including angular momentum loss through stellar winds, binary interactions, planetary system formation, and magnetic activity effects

C. Not all stars rotate, with key determinants including initial angular momentum from the parent molecular cloud, angular momentum loss during evolution, binary interactions, planetary system formation, and stellar radius fluctuations

D. Not all stars rotate, with key determinants including initial angular momentum from the parent molecular cloud, angular momentum loss during evolution, binary interactions, planetary system formation, and magnetic activity effects

E. All stars rotate, with key determinants including initial angular momentum from the parent molecular cloud, nuclear fusion efficiency, and binary interactions

F. Not all stars rotate, with key determinants including initial angular momentum from the parent molecular cloud, angular momentum loss during evolution, binary interactions, and magnetic activity effects

\textbf{Answer}: D
\end{tcolorbox}

\begin{tcolorbox}[breakable, title={Q\&A 4, Field: \textit{Computer Science}}]
\textbf{Question}: In a project management system, when trying to create an external dependency by linking a milestone nested within a phase to an external work package, a 'dependency to parent element exists' error message occurs. Both the phase and the work package are already dependent upon another milestone. The user finds it counterintuitive and questions potential circular dependencies, and notes that the dependency works when made from the external milestone directly to the phase. What is the appropriate solution to resolve this error while maintaining valid dependency constraints? 

\textbf{Answer}: The error occurs because the system prohibits creating a dependency from a child element (the milestone within the phase) to an external work package when the parent phase is already dependent on another milestone, to prevent indirect circular or redundant dependencies. The correct way to resolve this error is to create the dependency directly between the external milestone and the phase (the parent element), rather than linking the milestone nested inside the phase. By doing so, the dependency respects the system’s hierarchical constraints and avoids conflicts with existing parent-level dependencies.
\end{tcolorbox}

\section{Prompts of LLMs During Dataset Construction and Judgement.}
\label{app:llmprompt}
This benchmark is generated with the assistance of MLLMs and LLMs, and the relevant steps involved the following prompts.

\begin{tcolorbox}[breakable, title=Prompt for image to Q\&A]
You are a professional expert in data scraping and question generation. Your primary responsibility is to identify the content from forum post screenshots, accurately extract the original poster's question and others' replies, and then generate high-quality questions with a certain level of difficulty and misleading elements based on the scraped content.

Task Steps:

Step 1: Page Validity Check

1. Determine whether the current screenshot is a valid discussion page (not a 404 page, blank page, login page, or advertisement page)
2. If it is an invalid page, return directly: {"is\_valid": false}

Step 2: Question Validity Check

3. If the page is valid, check whether it contains a clear question

4. If no valid question is found (e.g., title only without content, meaningless characters, or simple informational posts without questions), return: {"is\_valid": false}

Step 3: Content Extraction and Processing

5. If there is a valid question and valid replies (substantial content, not just emojis or meaningless characters)

6. Extract the title, question description, and all relevant reply content to form a structured triplet: (question, answer, context)

Step 4: Result Generation

7. Based on the extracted core content, create high-quality academic questions that meet the following criteria:

- The question must be self-contained and independent of external context

- Rephrase the question using clear and natural language to enhance clarity, fluency, and understandability, avoiding ambiguity or vague expressions

- Be sufficiently professional — not simple knowledge quizzes or common-sense questions, but those requiring logical reasoning and analysis

- Require multi-step, cross-disciplinary complex reasoning

- Involve specialized or cutting-edge knowledge in specific domains, not purely general academic knowledge

- Avoid simple factual questions; emphasize comprehension and application

- The answer must be unique and verifiable, capable of being confirmed through reasoning or experimentation; exclude open-ended questions with multiple possible answers or solution paths

- Use the same language as the original forum post for the question and answer (e.g., if the post is in English, both question and answer should be in English)
8. Summarize and organize the most accurate, comprehensive, and detailed answer from the replies (leave blank if no valid reply exists)

9. Provide a detailed explanation and reasoning process for the answer (leave blank if no valid reply exists)

10. Set the "has\_answer" flag: 1 if there are valid replies and an answer can be extracted, 0 otherwise

Please strictly return the result in the following JSON format without adding any extra text:

\quad\{

\quad\quad``is\_valid": boolean,

\quad\quad``question": string,

\quad\quad``answer": string,

\quad\quad``explanation": string,

\quad\quad`has\_answer": integer

\quad\}
\end{tcolorbox}

\begin{tcolorbox}[breakable, title=Prompt for quality filter]
Task: Evaluate the quality of the following QA pair and retain only high-quality questions

Evaluation Criteria (all must be satisfied):

- Accuracy: All background knowledge, data, and facts involved in the question are accurate and error-free

- Completeness: The question provides all information necessary for solving it, with no missing or redundant information

- Professionalism: The question is sufficiently professional—not a simple knowledge quiz or common-sense question—but one that requires logical reasoning and analysis to solve

- Timelessness: The question does not involve time-sensitive content; it remains valid and true at all times

Difficulty Criteria: 
The question must meet the following requirements:

- Depth of Knowledge: Solving the question requires specialized or cutting-edge knowledge in a particular field

- Length of Reasoning Chain: The solution process involves multi-step, cross-disciplinary complex reasoning

- Abstraction and Synthesis Requirements: The question demands high-level abstract thinking and information synthesis

- Deceptiveness and Originality: The question requires creative or unconventional problem-solving approaches

QA pair to be evaluated:  

\quad Question: \{question\}  

\quad Answer: \{answer\}  

\quad Explanation: \{explanation\}  

Please return only the result in JSON format, including the following fields:  

- is\_qualified: boolean (whether the question meets the standards)  

- reason: string (reason for not meeting the standards; empty string if qualified)
\end{tcolorbox}

\begin{tcolorbox}[breakable, title=Prompt for multi-choice]
Task: Convert the question, answer, and explanation from the following QA pair into a high-quality multiple-choice question

Requirements:

1. Retain the core of the original question, but adjust the phrasing appropriately to suit the multiple-choice format

2. Generate 5 to 7 options, with exactly one correct answer and the rest as plausible distractors

3. Distractors should be misleading and belong to the same knowledge domain as the correct answer

4. Options must be randomly ordered

5. Ensure all options have a consistent grammatical structure

6. Maintain a high difficulty level, requiring multi-step reasoning to solve

7. The language used must be consistent with the original question—if the original QA pair is in English, the output must also be in English; do not change the language arbitrarily

QA pair:

\quad Question: \{question\}

\quad Answer: \{answer\}

\quad Explanation: \{explanation\}

Please return only the result in JSON format, including the following fields:

- question: string (the multiple-choice question)  

- options: array[string] (list of options, including one correct answer) 

- correct\_answer: string (the content of the correct answer)  

- explanation: string (explanation of why this option is correct)
\end{tcolorbox}

\begin{tcolorbox}[breakable, title=Prompt for short answer]
Task: Convert the question, answer, and explanation from the following QA pair into a high-quality short-answer question, and extract the core knowledge points

Requirements:

1. Retain the core of the original question, with appropriate rephrasing to better suit the short-answer format

2. Generate a detailed and comprehensive answer that includes all key information from the original explanation

3. The answer must be substantial, well-supported, and logically structured, avoiding vague or generic statements

4. Use natural, fluent, professional, and accurate language

5. Maintain a high difficulty level, reflecting professional knowledge and reasoning processes

6. Summarize 1–5 core knowledge points—these are the key concepts essential for determining the correctness of the answer

7. The language used must be consistent with the original question—if the original QA pair is in English, the output must also be in English; do not change the language arbitrarily

QA pair:

\quad Question: \{question\}  

\quad Answer: \{answer\}  

\quad Explanation: \{explanation\}  

Please return only the result in JSON format, including the following fields:

- question: string (the short-answer question)

- answer: string (detailed answer, incorporating the explanation)

- core\_knowledge\_points: array[string] (list of core knowledge points)
\end{tcolorbox}

\begin{tcolorbox}[breakable, title=Prompt of metric for answer evaluation]
From now on, your role is that of a professional grading teacher. Your task is to objectively score the <Student Answer> based on the <Reference Answer> and <Evaluation Points> that I provide. The main steps and rules include the following 6 points:

\begin{enumerate}
    \item Scoring Levels: There are only two levels, 0 and 1. No other scores are allowed, so please do not give any score other than 0 or 1.

    \item Since I have not provided you with the <Question>, you only need to judge whether the content of the <Student Answer> meets the <Evaluation Points>. Do not imagine or infer the content of the <Question>. Note that <Evaluation Points> take precedence over the <Reference Answer>. If there is a conflict between them, you must prioritize the requirements of the <Evaluation Points>.

    \item You need to first determine whether the <Student Answer> meets the requirements of the <Evaluation Points> (if no <Evaluation Points> are provided, then by default, the results and requirements in the <Reference Answer> are considered as the evaluation points). The answer must meet all conditions in the evaluation points. If any one of them is not satisfied, you should immediately give a score of 0, stop executing Step~4, and output the result according to the format in Step~5. If all conditions are met, continue to Step~4.

    \item Determine whether the final answer in the <Student Answer> is correct:
    \begin{itemize}
        \item You do not need to consider whether the process is correct, only the final answer.
        \item If the final answer is identical or synonymous to the <Reference Answer> (e.g., ``two times'' and ``2''), then give a score of 1. Otherwise, if it is inconsistent or not synonymous, give a score of 0.
        \item Some answers may contain errors, but if later self-corrected, the final result after correction should be regarded as the actual answer.
        \item Some answers may include self-analysis or feedback, possibly repeating outputs multiple times. Do not be misled by intermediate outputs---only use the last final result for grading.
    \end{itemize}

    \item Format Requirements: Before outputting your result, first check your analysis process and score to ensure they are reasonable and correct. If there are any errors or omissions, revise them in time. Finally, when you believe everything is correct, provide the output strictly in the following format:
    \begin{itemize}
        \item Grading Basis: (Concise explanation of the grading reason, less than 100 tokens)
        \item Score: x (output must be ``0'' or ``1'')
        \item JSON:
        
        \{"answer\_score": your score\}
        
    \end{itemize}

    \item Before you formally start the grading task, to improve your grading accuracy and understanding of the scoring standards, I will first provide 
    a simulated scoring example. By studying this example, you will become more familiar with the grading process and master the grading techniques:
\end{enumerate}

\subsection*{Example}
<Reference Answer>:
\begin{quote}
5. Move to the kitchen. \\
3. Find the kettle, check that it has water. \\
2. Pick up the kettle and move to the balcony. \\
4. Water the flowers. \\
Path: 5, 3, 2, 4
\end{quote}

<Evaluation Points>:
\begin{enumerate}
    \item The ``watering process'' in the student's answer must be exactly consistent with the reference answer; otherwise, score 0.
    \item The ``path'' in the student answer must be exactly consistent with the reference answer, otherwise score 0.
\end{enumerate}

<Student Answer>:
\begin{quote}
5. Move to the kitchen. \\
1. Find a sponge, check if it has water. \\
6. Pick up the sponge and move to the balcony. \\
4. Water the flowers. \\
Path: 5, 1, 6, 4
\end{quote}

Grading Basis: Both process and path are inconsistent with the reference answer, hence score 0.  

Score: 0 

JSON:

\{"answer\_score": 0\}

\bigskip
I hope you can fulfill the role of grading teacher. If you perform well, I will give you appropriate rewards. Please strictly follow the output format I provide; otherwise, I will penalize you. Also, always use the final result of the answer for grading, and do not be misled by intermediate outputs.
\end{tcolorbox}

\section{Forum list}
\label{app:forums}
The forums selected include, but are not limited to, the following:\\
\url{https://mathoverflow.net}/\\ 
\url{https://geant4-forum.web.cern.ch/}\\ 
\url{https://root-forum.cern.ch/}\\ 
\url{https://www.physicsforums.com/}\\ 
\url{https://www.biostars.org/}\\ 
\url{https://eaaforums.org/}\\ 
\url{http://www.mjtd.com/}\\ 
\url{https://www.practicalmachinist.com/}\\ 
\url{https://diysolarforum.com/}\\ 
\url{https://cr4.globalspec.com/thread/88025/High-Voltage-Engineering}\\ 
\url{https://www.elitetrader.com/}\\ 
\url{https://answers.justia.com/}\\ 
\url{https://www.lawanswers.com.au/forums/defamation-law-forum.25/}\\ 
\url{https://artofproblemsolving.com/community}\\ 
\url{https://board.asm32.info/}\\ 
\url{https://learn.microsoft.com/en-us/answers/topics/azure-digital-twins.html}\\ 
\url{https://www.biostars.org/}\\ 
\url{http://forums.corvetteforum.com/index.php}\\ 
\url{http://www.cplusplus.com/forum/}\\ 
\url{https://discuss.dvc.org/}\\ 
\url{https://forum.filezilla-project.org/index.php}\\ 
\url{http://www.fluka.org/fluka.php?id=mailinglist\&mm2=6}\\ 
\url{https://engx.theiet.org/}\\ 
\url{https://forums.majorgeeks.com/}\\ 
\url{https://community.myfitnesspal.com/en/categories/forums}\\ 
\url{http://www.openedv.com/}\\ 
\url{https://www.physicsforums.com/}\\ 
\url{https://discuss.pytorch.org/}\\ 
\url{https://forums.raspberrypi.com/}\\ 
\url{https://www.reddit.com/r/math/}\\ 
\url{https://root-forum.cern.ch/}\\ 
\url{https://community.spiceworks.com/}\\ 
\url{https://stackoverflow.com/questions/tagged/robotics}\\
\url{https://www.statalist.org/forums/forum/general-stata-discussion/general}\\ 
\url{https://www.techpowerup.com/forums/}\\ 
\url{https://pcengine.proboards.com/}\\ 
\url{https://www.vnpy.com/forum/}\\ 
\url{http://forums.vwvortex.com/}\\ 
\url{https://guba.eastmoney.com/}\\ 
\url{https://bbs.pinggu.org/}\\ 
\url{http://www.mjtd.com/}\\ 
\url{http://www.3dportal.cn/}\\ 
\url{http://www.proewildfire.cn/}\\ 
\url{https://www.armbbs.cn/}

\textbf{All these forums are publicly accessible platforms that either permit or do not prohibit the use of their content for the AI community.}

\section{Expert Qualification and Rules}\label{appsec:experts_qualification}
A total of 209 experts participated in this project, with their qualifications categorized and described across three dimensions: (1) professional field, (2) educational background, and (3) institutional affiliation. All data was self-reported by the experts and verified internally.

\textbf{1. Professional Domain Distribution}. Experts are categorized according to their self-declared primary field. To maintain data integrity, all original categories—including non-academic ones—are retained, totaling 209 individuals.

\begin{table}[h]
\centering
\begin{tabular}{lcc}
\toprule
\textbf{Primary Domain} & \textbf{Count} & \textbf{Pct. (\%)} \\ \midrule
CS \& Mathematical Sciences & 87 & 41.6 \\
STEM & 60 & 28.7 \\
Business & 20 & 9.6 \\
Medical & 13 & 6.2 \\
Law & 9 & 4.3 \\
Humanities \& Social Sciences & 8 & 3.8 \\
Lifestyle \& Hobbies & 5 & 2.4 \\
Education & 4 & 1.9 \\
Languages & 4 & 1.9 \\
Creative Writing & 2 & 1.0 \\
\midrule
\textbf{Total} & \textbf{209} & \textbf{100.0} \\
\bottomrule
\end{tabular}
\caption{Distribution of Experts by Primary Professional Domain}
\label{tab:domain}
\end{table}

\textbf{2. Educational Attainment}.
The educational background of the expert panel is highly advanced, with the vast majority holding graduate degrees.

\begin{table}[h]
\centering
\begin{tabular}{lcc}
\toprule
\textbf{Degree Level} & \textbf{Count} & \textbf{Pct. (\%)} \\ \midrule
Ph.D. & 94 & 45.0 \\
Master’s & 97 & 46.4 \\
Bachelor’s & 15 & 7.2 \\
Not specified / Other & 3 & 1.4 \\
\midrule
\textbf{Total} & \textbf{209} & \textbf{100.0} \\
\bottomrule
\end{tabular}
\caption{Educational Attainment of Experts}
\label{tab:education}
\end{table}

\textbf{3. Institutional Affiliation}.
The institutional prestige is classified using the QS World University Rankings and inclusion in national elite research systems. Institutions are grouped as Table~\ref{tab:institution}.

The National-Level Research Institutes include the University of Chinese Academy of Sciences, Chinese Academy of Medical Sciences, People’s Liberation Army General Hospital, etc., which entities with doctoral granting authority and national research mandates.

During the data construction process, the experts are asked to select forums that  either permit or do not prohibit the
use of their content for the AI community. 
All experts are paid reasonable remuneration based on their workload in accordance with the market price.
\begin{table}[h]
\centering
\begin{tabular}{lcc}
\toprule
\textbf{Institution Category} & \textbf{Count} & \textbf{Pct. (\%)} \\\midrule
QS World Top 50 Univ. & 91 & 43.5 \\
QS World Rank 51–200 Univ. & 32 & 15.3 \\
National-Level Research Inst. & 54 & 25.8 \\
Other/Unspecified Inst. & 32 & 15.3 \\
\midrule
\textbf{Total} & \textbf{209} & \textbf{100.0} \\
\bottomrule
\end{tabular}
\caption{Institutional Affiliation of Experts}
\label{tab:institution}
\end{table}

\section{Rules for preliminary-experiment driven difficulty categorize for \name}\label{appsec:difficulty_distribution}
To divide the the tasks in \name into different difficulty levels, we randomly select 6-8 SOTA LLMs to evaluate whether it could solve it correctly and determine its difficulty level by the proportion of models solving it. The number of different level tasks and corresponding criterion is shown in Table~\ref{tab:difficulty}. 

\begin{table}[h]
\centering
\begin{tabular}{lcc}
\toprule
\textbf{Difficulty Level} & \textbf{Pass Rate Criterion} & \textbf{Nums} \\
\midrule
L1 (Easy)   & $>50\%$        & 137 \\
L2 (Medium) & $30\%-50\%$    & 133 \\
L3 (Hard)   & $<30\%$        & 160 \\
\midrule
\textbf{Overall}     & --             & \textbf{430} \\
\bottomrule
\end{tabular}
\caption{Difficulty level distribution of \name. The level of each task is calculated by the average pass rate of all models evaluated during the preliminary experiment.}
\label{tab:difficulty}
\end{table}

\section{Detailed Error Type Distribution on Different Primary Domains in \name.}
\label{app:primary error attribute}
The detailed distributions of error attributions of different primary domains in \name is shown in Table~\ref{tab:error-by-discipline}.

\begin{table*}[h]
\centering
\small
\setlength{\tabcolsep}{5pt}
\begin{tabular}{lccccccc}
\toprule
\textbf{Discipline} 
& \textbf{Over-simplify} 
& \textbf{Reasoning} 
& \textbf{Knowledge} 
& \textbf{Hallucination} 
& \textbf{Misinterpretation} 
& \textbf{Calculation} 
& \textbf{Others} \\
\midrule
Natural Sci   & 6  & 20 & 50 & 4 & 18 & 2 & 0 \\
Applied Tech  & 8  & 12 & 44 & 4 & 30 & 0 & 2 \\
Maths         & 4  & 10 & 40 & 0 & 34 & 6 & 6 \\
Life Sci      & 4  & 24 & 52 & 0 & 18 & 0 & 2 \\
EE            & 6  & 18 & 40 & 2 & 34 & 0 & 0 \\
Social Sci    & 6  & 22 & 42 & 2 & 26 & 0 & 2 \\
CS            & 4  & 8  & 58 & 0 & 28 & 0 & 2 \\
\bottomrule
\end{tabular}
\caption{Percentage distribution (\%) of error types across disciplines in \name. 
Each discipline contains 50 questions; percentages are obtained by multiplying raw counts by 2.}
\label{tab:error-by-discipline}
\end{table*}

\section{Detailed Scores of LLMs on MMLU-pro and HLE}
\label{app:scoreonotherbench}
The detailed scores and rankings of different models on \name, MMLU-pro and HLE are shown in Table~\ref{tab:benchmark_comparison}.

\begin{table*}[h]
\centering
\small
\setlength{\tabcolsep}{6pt}
\begin{tabular}{l cc cc cc}
\toprule
\textbf{Model} 
& \multicolumn{2}{c}{\textbf{\name}} 
& \multicolumn{2}{c}{\textbf{MMLU-pro}} 
& \multicolumn{2}{c}{\textbf{HLE}} \\
\cmidrule(lr){2-3} \cmidrule(lr){4-5} \cmidrule(lr){6-7}
& Score & Rank & Score & Rank & Score & Rank \\
\midrule
GLM-4.6                     & 50.5 & 3 & 83.2 & 7 & 17.2 & 6 \\
Gemini-3-Pro               & 51.2 & 2 & 90.2 & 1 & 37.7 & 1 \\
Kimi-K2-Thinking& 47.9 & 6 & 84.8 & 5 & 23.9 & 4 \\
GPT-5-High& 54.4 & 1 & 87.1 & 3 & 26.3 & 2 \\
Claude-Sonnet-4.5-Thinking& 49.5 & 4 & 87.5 & 2 & 13.7 & 7 \\
DeepSeek-V3.2-Thinking& 49.5 & 4 & 85.0 & 4 & 25.1 & 3 \\
Qwen3-235B-Thinking& 47.6 & 7 & 84.5 & 6 & 18.2 & 5 \\
\bottomrule
\end{tabular}
\caption{Performance comparison of LLMs across \name, MMLU-pro, and HLE. The scores of models on MMLU-pro and HLE are mainly obtained from models' official technical reports and model cards.}
\label{tab:benchmark_comparison}
\end{table*}

\section{Case Studies on Model Performance when with/without Tools}
\label{app:caseontool}

\subsection{CI Tool}

\begin{tcolorbox}[
  breakable,
  title={ Typical Failure Case: \name JupyterCI Leads to Decreased Model Score}
]

\textbf{Task Description}

\textbf{Question.}
Can every real number in the unit interval $[0,1]$ be expressed as the product of finitely many elements of the Cantor set?
If so, what is the minimum number of Cantor set elements required?

\begin{itemize}
\item[A.] Yes; the minimum number required is 2
\item[B.] Yes; the minimum number required is 4
\item[C.] Yes; the minimum number required is 3
\item[D.] Yes; the minimum number required is 1
\item[E.] Yes; the minimum number required is 5
\item[F.] No; there exist real numbers in $[0,1]$ that cannot be expressed as the product of finitely many Cantor set elements
\end{itemize}

\vspace{0.5em}
\textbf{Ground-truth answer:}
Yes. The minimum number is 3.

\vspace{0.8em}
\textbf{Model Behavior without JupyterCI}

Without external tools, the model directly recalls a known theoretical result in fractal geometry and real analysis:
\[
C \cdot C \cdot C = [0,1],
\]
while the double product $(C \cdot C)$ does not cover the entire interval.
The model therefore selects Option C (3 elements) correctly.
This demonstrates that the model possesses the relevant latent mathematical knowledge and can apply it via pure language-based reasoning.

\vspace{0.8em}
\textbf{Model Behavior with JupyterCI}

\textbf{Step 1: Tool Invocation and Problem Reformulation}

Upon being equipped with a Code Interpreter (CI), the model reformulates the original question as a computational coverage verification task.
Specifically, it:
\begin{itemize}
\item Constructs finite-level approximations $(C_n)$ of the Cantor set;
\item Computes the pairwise product $(C_n \cdot C_n)$;
\item Evaluates numerical statistics such as maximum gap, nearest distance to grid points, and empirical coverage over $[0,1]$.
\end{itemize}

\textbf{Step 2: Interpretation of Numerical Outputs}

The CI outputs indicate persistent gaps and non-zero distances for all tested finite levels $(n \leq 10)$, for example stable maximum gaps around $(0.11)$.
The model interprets these results as evidence that the product of two Cantor elements fails to cover $[0,1]$.

\textbf{Step 3: Tool-Induced Reasoning Lock-in}

After committing to the computational path, the model does not reconsider or override the numerical evidence with known theory.
Instead, it:
\begin{itemize}
\item Constructs a post-hoc theoretical justification (via logarithmic transformation and thickness arguments);
\item Incorrectly concludes that two factors suffice, selecting Option A.
\end{itemize}

\vspace{0.8em}
\textbf{Failure Analysis}

With JupyterCI enabled, the model selects Option A (2 elements), yielding an incorrect answer, despite answering correctly in the tool-free setting.

\textbf{Summary of Failure Mode}

This case exemplifies a systematic degradation pattern induced by external tools:
\begin{itemize}
\item Tools bias the model toward procedural, simulation-based reasoning, even when inappropriate;
\item Finite numerical evidence is misused to answer questions about infinite mathematical objects.
\end{itemize}

\vspace{0.8em}
\textbf{Takeaway}

The resulting reasoning lock-in prevents recovery via pure language-based knowledge.
Importantly, the error arises not from missing information or computational failure, but from a mismatch between the tool’s affordances and the task’s epistemic structure.

\end{tcolorbox}

\subsection{Search Tool}

\begin{tcolorbox}[
  breakable,
  title={Typical Failure Case: \name Search Tool Leads to Decreased Model Score}
]

\textbf{Question:}
In a computational study of perovskite solar cell materials, a research team performs an excited-state geometry optimization on the
CH$_3$NH$_3$PbI$_3$ crystal.
The material crystallizes in a tetragonal lattice.
The team first constructs a $2 \times 2 \times 1$ supercell and maps the primitive-cell coordinates to the supercell.
After building the structure, they carry out an $S_1$ excited-state optimization using the ORCA software package at the PBE0/def2-SVP level of theory.
Upon convergence with the \texttt{VeryTightOpt} criterion, a single significant imaginary frequency of
$-32\,\mathrm{cm^{-1}}$ is observed.

Which of the following statements regarding this study is correct?

\begin{itemize}
\item[A.] The mapping from the primitive cell to the $2 \times 2 \times 1$ supercell fails because the axial-angle characteristics of the tetragonal lattice prevent the use of simple translation vectors.
\item[B.] To resolve the imaginary frequency, one should immediately search for a transition state using the \texttt{Opt=TS} keyword in Gaussian.
\item[C.] Errors in the supercell coordinate construction have a direct causal relationship with the appearance of the imaginary frequency in the excited-state optimization.
\item[D.] The most reasonable way to handle the imaginary frequency is to apply ORCA’s iterative optimization method by including the \texttt{\%compound iterativeOptimization.cmp} setting.
\item[E.] Owing to the high symmetry of the tetragonal lattice, the coordinates of the $2 \times 2 \times 1$ supercell can be obtained from the primitive-cell coordinates via simple matrix multiplication.
\end{itemize}

\vspace{0.5em}
\textbf{Ground-truth answer:}
E — Supercell coordinates in a tetragonal lattice can be obtained via a simple linear (matrix) transformation from the primitive cell.

\vspace{0.8em}
\textbf{Model Behavior without Search}

When no external search tool is available, the model selects \textbf{E}. Its reasoning focuses on the crystallographic principle that supercell construction is a linear transformation independent of vibrational artifacts arising in excited-state optimization. The imaginary frequency is correctly treated as irrelevant to the coordinate-mapping question posed. 

\vspace{0.8em}
\textbf{Model Behavior with Search}

When equipped with a search tool, the model invokes external retrieval and obtains ORCA manual pages and forum discussions describing the \textit{iterative optimization} compound script for removing imaginary frequencies. After incorporating this retrieved information, the model selects \textbf{D and E}, incorrectly elevating \textbf{D} as a correct choice. In particular, the model treats the existence of an ORCA workflow for eliminating imaginary frequencies as evidence that it is the most reasonable action in this scenario, despite the question not asking how to resolve the frequency issue.

\vspace{0.8em}
\textbf{Failure Analysis}

This case exemplifies a \textbf{search-induced failure mode} in \name, where tool usage degrades performance relative to the tool-free setting. First, the retrieved content is \textbf{semantically misaligned with the task objective}. The search tool returns method-level and documentation-level descriptions of how to remove imaginary frequencies, whereas the question requires a contextual judgment about crystallographic supercell mapping. Second, the model \textbf{misuses existence as normativity}. The availability of a technically valid ORCA procedure is incorrectly promoted to a normative recommendation for this specific problem, without considering whether the procedure is relevant to what is being asked. Third, \textbf{external retrieval disrupts internal domain reasoning}. Exposure to detailed procedural information diverts the model’s attention away from its original, correct crystallographic reasoning path. Finally, tool availability induces an \textbf{over-engineering bias}: instead of answering a conceptual multiple-choice question, the model proposes an executable computational workflow, leading it to select an option that is technically plausible but conceptually out of scope.

\vspace{0.8em}
\textbf{Takeaway}

This example demonstrates that, for low-retrievability professional questions, external search can introduce technically correct but contextually inappropriate information.

\end{tcolorbox}

\end{document}